\documentclass[lettersize,journal]{IEEEtran}
\usepackage{amsmath,amsfonts}
\usepackage{algorithmic}
\usepackage{algorithm}
\usepackage{array}
\usepackage[caption=false,font=normalsize,labelfont=sf,textfont=sf]{subfig}
\usepackage{textcomp}
\usepackage{stfloats}
\usepackage{url}
\usepackage{multirow}
\usepackage{verbatim}
\usepackage{graphicx}
\usepackage{booktabs}
\usepackage{cite}
\usepackage{pifont}
\usepackage{color}
\usepackage{hyperref}
\hypersetup{
 colorlinks=true,
 linkcolor=blue,
 anchorcolor=blue,
 citecolor=green
}

\usepackage{etoolbox}
\makeatletter
\patchcmd{\@makecaption}
{\scshape}
{}
{}
{}
\makeatother

\makeatother

\begin{document}

\title{Unveiling Context-Related Anomalies: Knowledge Graph Empowered Decoupling of Scene and Action for Human-Related Video Anomaly Detection}


\author{\IEEEauthorblockN{Chenglizhao Chen~~~~Xinyu Liu~~~~Mengke Song$^{\dag}$~~~~Luming Li~~~~Xu Yu~~~~Shanchen Pang\\
		\IEEEauthorblockA{College of Computer Science and Technology, China University of Petroleum (East China)\\
			\vspace{-0.5cm}
			\thanks{\dag\ Corresponding author: Mengke Song (songsook@163.com)}
}}}

\markboth{IEEE Transactions on Multimedia, VOL.XX, NO.XX, XXX.XXXX}%
{Shell \MakeLowercase{\textit{et al.}}: Bare Demo of IEEEtran.cls for Journals}

\maketitle
\begin{abstract}	
	Detecting anomalies in human-related videos is crucial for surveillance applications. Current methods primarily include appearance-based and action-based techniques.
	Appearance-based methods rely on low-level visual features such as color, texture, and shape. They learn a large number of pixel patterns and features related to known scenes during training, making them effective in detecting anomalies within these familiar contexts. However, when encountering new or significantly changed scenes, \textbf{\textit{i.e.}}, unknown scenes, they often fail because existing SOTA methods do not effectively capture the relationship between actions and their surrounding scenes, resulting in low generalization.
	In contrast, action-based methods focus on detecting anomalies in human actions but are usually less informative because they tend to overlook the relationship between actions and their scenes, leading to incorrect detection. For instance, the normal event of running on the beach and the abnormal event of running on the street might both be considered normal due to the lack of scene information.
	In short, current methods struggle to integrate low-level visual and high-level action features, leading to poor anomaly detection in varied and complex scenes.
	To address this challenge, we propose a novel decoupling-based architecture for human-related video anomaly detection (DecoAD). DecoAD significantly improves the integration of visual and action features through the decoupling and interweaving of scenes and actions, thereby enabling a more intuitive and accurate understanding of complex behaviors and scenes.
	DecoAD supports fully supervised, weakly supervised, and unsupervised settings. In the UBnormal dataset, DecoAD increases the AUC by 1.1\%, 3.1\%, and 1.7\% in fully supervised, weakly supervised, and unsupervised settings, respectively. In the NWPU Campus dataset, it increases the AUC by 0.2\% in both weakly supervised and unsupervised settings. We make our source code and datasets publicly accessible at \url{https://github.com/liuxy3366/DecoAD}.

\end{abstract}

\section{Introduction}
\label{sec:intro}
Video anomaly detection is a critical task that involves identifying unusual or abnormal events, behaviors, and activities within video sequences.
This task is essential in several domains, including security, surveillance, public safety, and abnormal behavior analysis~\cite{8767943,9263356,10174739}.
Human-related video anomaly detection refers to specifically detecting anomalies involving human subjects.
This branch of anomaly detection primarily focuses on identifying abnormal activities such as criminal behavior, accidents, or unusual behavior patterns displayed by individuals.
The traditional methods include appearance-based methods and action-based methods.
\begin{figure}[!t]
	\centering{\includegraphics[width=1\linewidth]{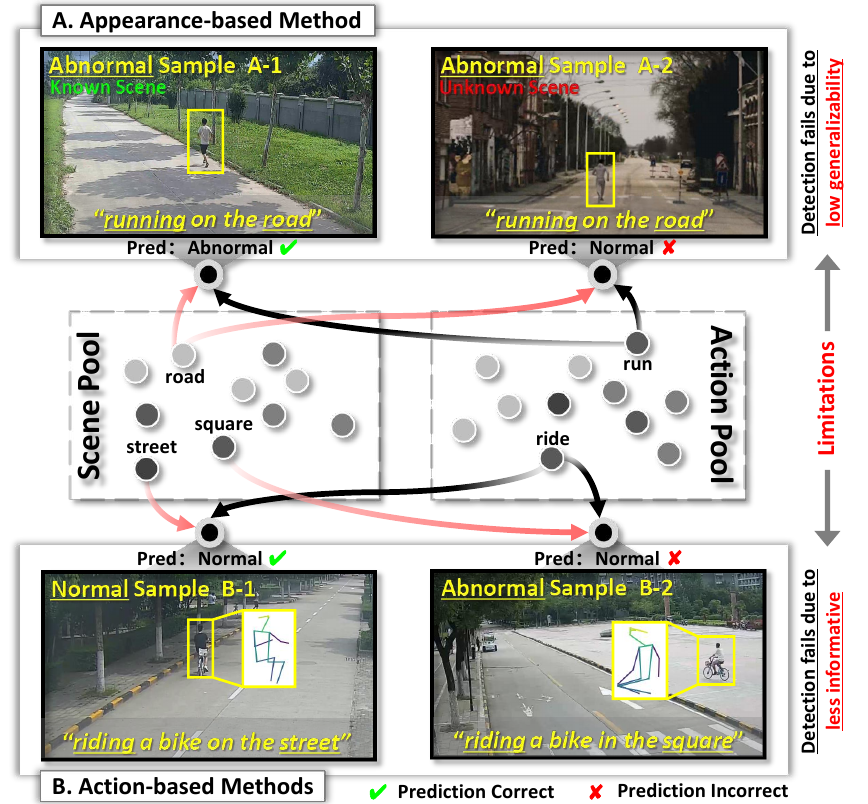}}
	\vspace{-0.4cm}
	\caption{Reveal the limitations of existing methods: appearance-based methods fail to detect anomalies due to their low generalizability (\textbf{A}),  action-based methods fail due to their less informative (\textbf{B}). ``Known Scene" refers to the scene present in the training set, and ``Unknown Scene" refers to the scene not present in the training set or those that have significant changes.
	}
	\label{fig:demo}
	\vspace{-0.45cm}
\end{figure}

Most video anomaly detection methods rely on low-level visual features, namely appearance-based methods, to capture human behavior~\cite{zhou2020skeleton,li2022human}. These methods learn to recognize extensive pixel patterns and features related to known scenes during training, thus enabling effective anomaly detection within these familiar contexts. 
However, because these methods rely solely on low-level visual features such as color, texture, and shape, they fail to effectively capture the relationship between actions and their surrounding scenes. This results in low generalization and high sensitivity to factors that significantly alter the visual appearance of objects, such as changes in lighting conditions, camera viewpoints, and object occlusion~\cite{sabih2022novel,panariello2022consistency,10504300}.
Consequently, their performance significantly degrades when encountering new or significantly changed scenes. For instance, as shown in Fig.~\ref{fig:demo}-A, appearance-based methods can successfully detect a running person in a known road scene but may fail in an unknown scene. To overcome this limitation, many existing video anomaly detection methods consider using high-level action features.

Methods using high-level action features can be categorized as action-based methods. These methods utilize high-level features extracted from videos during training, such as skeletal data and pose estimation~\cite{yu2023regularity,mishra2024skeletal}. These features are compact, well-structured, and highly descriptive of human behaviors and actions, thereby significantly enhancing the model's generalizability. However, existing methods primarily focus on identifying anomalies in human actions, such as running or fighting~\cite{boekhoudt2021hr,li2023human,he2024compressed,jiang2023physically}. These methods are often less informative because they tend to overlook the relationship between scenes and human actions. For example, as shown in Fig.~\ref{fig:demo}-B, existing action-based methods cannot distinguish between riding a bicycle on the street and riding it in a square. This lack of contextual information leads to detection failures.

\begin{figure*}[!t]
	\centering{\includegraphics[width=1\linewidth]{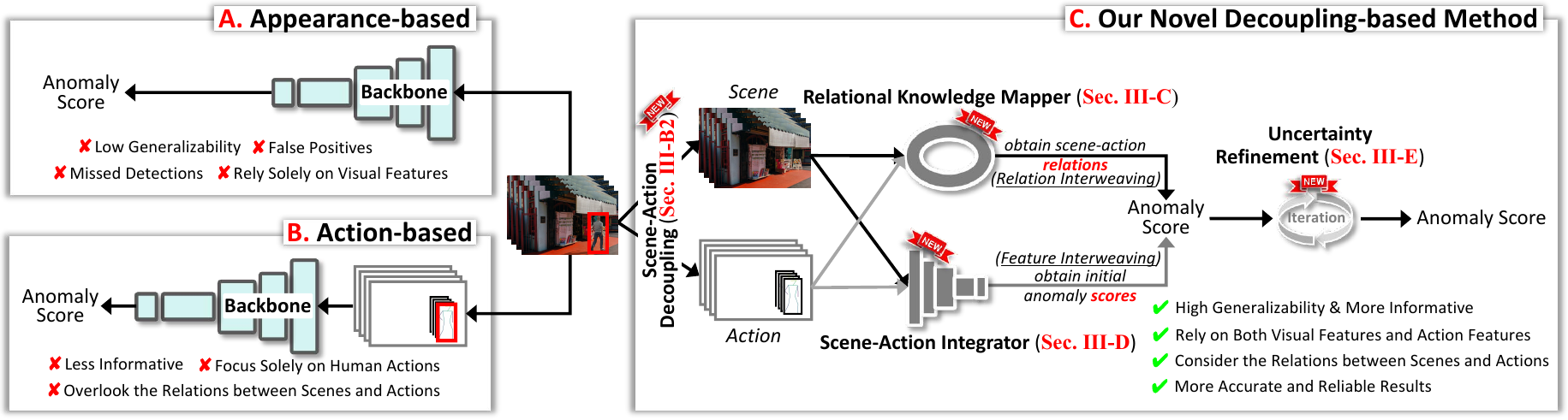}}
	\vspace{-0.4cm}
	\caption{Compared to appearance-based methods (\textbf{A}), which only rely on low-level visual features, and action-based methods (\textbf{B}), which ignore the relationship between scenes and human actions, our decoupling-based method (\textbf{C}) introduces the concept of ``Scene-Action Interweaving". Fully considering the complex connections between actions and the surrounding environment in different video clips.}
	\label{fig:motivation}
	\vspace{-0.45cm}
\end{figure*}

Whether appearance-based or action-based, the methods almost always use implicit associations through the model's internal learning mechanisms to capture and represent the relationships between data, as shown in Fig.~\ref{fig:motivation}-A, B. However, using implicit associations makes it challenging to effectively capture the relationships between features, leading to somewhat chaotic handling of these relationships.
Additionally, these methods tend to memorize training data, meaning the models can only detect anomalies or actions that appeared in the training set. When new scenes or anomaly events occur, the models need to be retrained, which lacks generalizability. In practical applications, companies often do not have sufficient computational resources to retrain models, so they can only use pre-trained models directly. Therefore, a method balancing performance and generalizability is urgently needed.


To further enhance the performance and generalizability of the model, this study introduces a novel \textbf{deco}upling-based architecture for human-related video \textbf{a}nomaly \textbf{d}etection (DecoAD). 
DecoAD uses explicit associations by fusing visual and action features to compensate for the limitations of low-level visual features and address the issue of being less informative.
DecoAD introduces the concept of ``Scene-Action Interweaving", which decouples scenes and human actions within video clips and interweaves them with elements from other clips. This approach aims to explore and understand the complex relationships between these scenes and actions.
Specifically, ``Scene-Action Interweaving" consists of two main parts: ``Relation Interweaving" and ``Feature Interweaving". ``Relation Interweaving" focuses on learning deep and complex relational patterns between scenes and human actions. ``Feature Interweaving" aims to comprehensively understand complex, context-related, and interrelated patterns.


To achieve ``Scene-Action Interweaving'', we have designed four main components, as illustrated in Fig.~\ref{fig:motivation}-C: Scene-Action Decoupling (Sec.~\ref{sec:decoup}), Relational Knowledge Mapper (Sec.~\ref{sec:rkm}), Scene-Action Integrator (Sec.~\ref{sec:sad}), and Uncertainty Refinement (Sec.~\ref{sec:uit}).
Firstly, we decouple scenes and associated human action elements within video clips. 
Then, the Relational Knowledge Mapper performs ``Relation Interweaving" to obtain scene-action relations.
This involves intricately interweaving the relations of scenes and human actions from different video clips, aiming to understand their complex interactions.
Next, the Scene-Action Integrator is used for ``Feature Interweaving" to obtain initial anomaly scores, representing the likelihood of anomalies in the video clips. Finally, Uncertainty Refinement ensures that video clips predicted with uncertain anomaly scores are iteratively fed into the Scene-Action Integrator to obtain more accurate results.

DecoAD has been trained under fully/weakly-supervised and unsupervised conditions, outperforming existing human-related video anomaly detection methods on three widely-used benchmark datasets --- NWPU Campus~\cite{cao2023new}, UBnormal~\cite{acsintoae2022ubnormal}, and HR-ShanghaiTech~\cite{liu2018future}.The main contributions of this work are then summarized as following.


\begin{itemize}
	\item
	In video anomaly detection tasks, the relationship between scenes and actions is often overlooked, leading to suboptimal detection performance. To address this, we propose a novel video anomaly detection framework, DecoAD, which emphasizes the relationship between scenes and actions, achieving finer-grained anomaly detection.
	\item
	Current approaches often mix action information with scene data, introducing noise and complexity. Our proposed Scene-Action Decoupling technique effectively separates scenes from actions and removes action information from scenes, minimizing noise and irrelevant features. This significantly boosts model generalization and ensures more reliable and precise anomaly detection.
	\item
	Existing methods primarily use implicit associations, which often overlook complex contextual information. We designed a Relational Knowledge Mapper that uses knowledge graphs to explicitly define the relationships between scenes and actions, improving anomaly detection accuracy and adapting to new data. We also developed a Scene-Action Integrator to combine scenes and actions for initial anomaly scores, and Uncertainty Refinement to iteratively refine scores for uncertain cases, enhancing detection reliability and accuracy across varied scenarios.
	\item
	We conduct detailed experiments on three widely used datasets, demonstrating that our method surpasses existing methods in both accuracy and robustness.
\end{itemize}

\section{Related Works}

\subsection{Video Anomaly Detection}
Video anomaly detection has long been a challenging task in the field of computer vision. Early research regarded it as an unsupervised learning task, more precisely, an out-of-distribution task, where the training process only involved normal samples~\cite{zhao2017spatio,8892741}. However, these early methods mostly rely on manually crafted features and statistical models, often resulting in limited generalization and robustness. With the advancement of deep learning technology~\cite{chen2021depth,song2023rethinking}, a wide array of new unsupervised learning methods have emerged in recent years~\cite{9055131,10539327,lin2022causal}. These methods aim to more effectively learn normal behavior patterns in video content. Due to the difficulty in annotating abnormal video data, unsupervised video anomaly detection has received widespread research attention. However, it is challenging to cover all normal samples during the training phase, often leading to higher false positive rates. To address this challenge, researchers have proposed weakly supervised video anomaly detection methods~\cite{zhang2019temporal,10330089,zhou2023dual,9540293,10471334}, primarily relying on the multiple instance learning framework to compensate for the absence of video-level labels. By striking a balance between annotation costs and detection performance, weakly supervised methods have shown considerable effectiveness. As research progresses, some datasets~\cite{acsintoae2022ubnormal} have begun to provide frame-level annotations, opening up new possibilities for fully supervised video anomaly detection~\cite{hirschorn2023normalizing}, and allowing existing fully supervised models to achieve higher detection accuracy.

In response to the diverse application demands of video data, we propose a novel video anomaly detection method that is flexible and applicable to unsupervised, weakly supervised, and even fully supervised learning scenarios.

\subsection{Human-Related Video Anomaly Detection}
Detecting anomalies in human-related videos is particularly challenging due to the complexity and diversity of human actions. Most human-related video anomaly detection methods fall into the category of appearance-based approaches~\cite{sun2023long}. Although these representations are simple and straightforward, they rely solely on low-level visual features such as color, texture, and shape to identify anomalies. This results in low generalizability of the models, and they often fail to detect anomalies when encountering new or significantly changed scenes.
In recent years, innovative advancements have been made in video anomaly detection of human behavior using action-based methods~\cite{flaborea2023multimodal,sato2023prompt}.
These methods leverage deep learning techniques to analyze the skeleton data extracted from videos to detect abnormal behavior. Using skeleton data as training data can mitigate or reduce the risk of privacy breaches. Additionally, human pose data can effectively reduce interference from noise and lighting factors. However, solely considering less informative skeletons without taking the scene into account can lead to critical issues. For example, the same action, such as a long jump, can be considered a normal event on a beach but an abnormal event on a road. This situation is common, where actions like running, dancing, or boxing can have different effects in different scenes.

\subsection{Knowledge Graph}
Knowledge graph is a complex graph-like data structure that organizes and represents knowledge to reveal relationships and connections between data~\cite{paulheim2017knowledge,ji2021survey}. It is widely applied in various fields, such as search engine optimization, recommendation systems, natural language processing, and social network analysis. Knowledge graphs effectively integrate and correlate vast amounts of information in these applications, providing users with more accurate and insightful results.

Our research work introduces a pioneering application of knowledge graphs in the field of video anomaly detection. In our approach, we decompose the video content into action and background elements and then utilize the knowledge graph to describe and understand the relationships between these elements. Within the knowledge graph, the relationships between scenes and actions are annotated as ``normal" or ``abnormal", offering an intuitive understanding and explanation of abnormal behaviors for the model.

\begin{figure*}[!t]
	\centering{\includegraphics[width=1\linewidth]{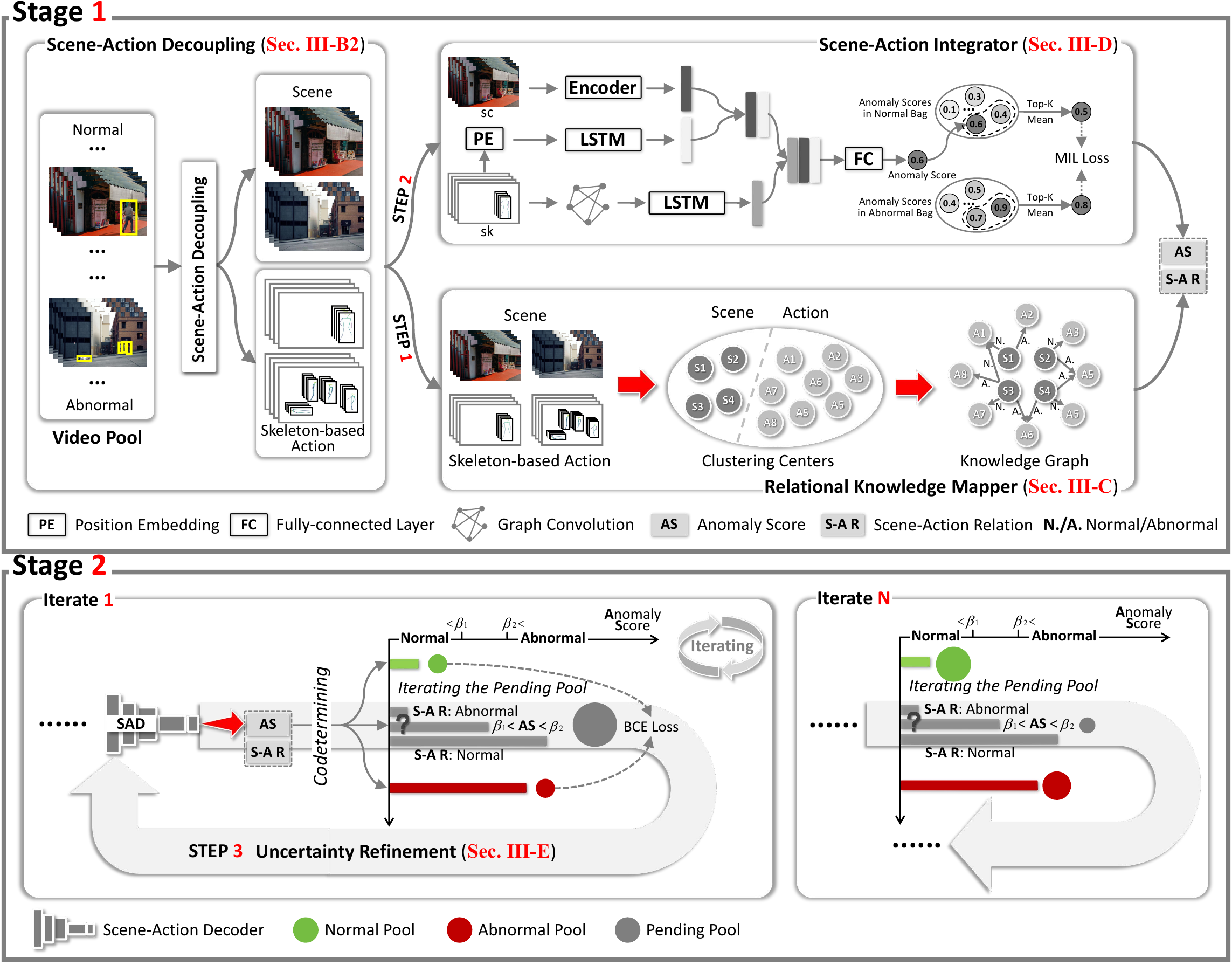}}
	\vspace{-0.5cm}
	\caption{Pipeline of the proposed DecoAD. DecoAD consists of three steps --- Step1: Relational Knowledge Mapper (RKM), Step2: Scene-Action Integrator (SAI) (\textbf{Stage 1}) and Step3: Uncertainty Refinement (\textbf{Stage 2}).}
	\label{fig:pipeline}
	\vspace{-0.45cm}
\end{figure*}

\section{Proposed Method}
\subsection{Method Overview}
Our proposed method, DecoAD, as illustrated in Fig.~\ref{fig:pipeline}, consists of four main components: Scene-Action Decoupling (Sec.~\ref{sec:decoup}), Relational Knowledge Mapper (Sec.~\ref{sec:rkm}), Scene-Action Integrator (Sec.~\ref{sec:sad}), and Uncertainty Refinement (Sec.~\ref{sec:uit}).

In \textbf{Stage 1}, we begin by decoupling a video clip into scenes and their associated skeleton-based human actions. Next, in Step1, we employ the Relational Knowledge Mapper to interweave these actions and scenes with those from different video clips. This involves constructing a detailed knowledge graph that captures the relationships between the scenes and skeleton-based actions, resulting in scene-action relations.
In Step2, the Scene-Action Integrator is utilized to generate initial anomaly scores. These scores indicate the likelihood of anomalies present in the video clips. Finally, in \textbf{Stage 2}, we incorporate Uncertainty Refinement (Step3) to ensure the Scene-Action Integrator iteratively processes video clips that are predicted with uncertain anomaly scores. This iterative process helps to obtain more accurate results. 
It is worth noting that this paradigm is trained using both fully-supervised and weakly-supervised approaches, while unsupervised methods do not undergo iterative training.



\subsection{Preliminaries}
\subsubsection{Scene-Action Interweaving}
Building on existing human-related video anomaly detection methods~\cite{sultani2018real,hirschorn2023normalizing}, it is essential to emphasize integrating scene context with human actions for more effective anomaly detection. Current approaches, whether appearance-based~\cite{tian2021weakly,chen2023mgfn} or action-based~\cite{yan2018spatial,cheng2020skeleton}, can recognize abnormal human actions like running or fighting. However, they frequently fail to consider the context of the scenes and actions, which can be crucial for accurately identifying context-related anomalies.

Thus, as mentioned in Sec.~\ref{sec:intro}, we propose the concept of ``Scene-Action Interweaving" for the first time. By decoupling scenes and human actions in video clips and interweaving them with elements from other video clips, we explore and understand the complex relationships and interactions between these scenes and actions. By combining and analyzing diverse elements from different video clips, we form a comprehensive semantic network, thereby enhancing the detection of context-related anomalies.



\subsubsection{Scene-Action Decoupling}
\label{sec:decoup}
\begin{figure}[!t]
	\hspace{-0.22cm}
	\centering{\includegraphics[width=1\linewidth]{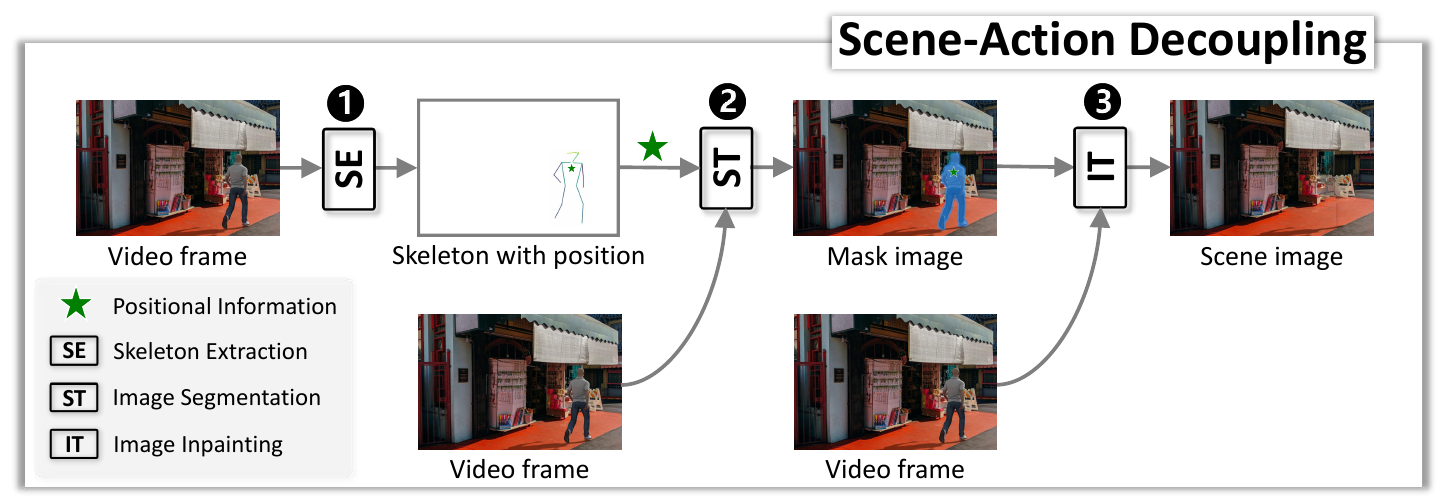}}
	\vspace{-0.2cm}
	\caption{Pipeline for processing image in Scene-Action Decoupling.}
	\label{fig:iam}
	\vspace{-0.45cm}
\end{figure}

The core concept of ``Scene-Action Interweaving” involves exploring the complex relationships between scene contexts and human actions by integrating them with another video clip to capture comprehensive interactions.
To facilitate this, we first decouple  scenes and their associated human actions within each video clip.
For the extraction of human actions, we employ a human skeleton extraction tool, similar to the methods used in existing human-related video anomaly detection research~\cite{sultani2018real,hirschorn2023normalizing}. Specifically, we derive skeletal data $a$ from the video clip ${\rm V}$ as a representation of actions\footnote{In this study, we treat skeletal data as equivalent to actions, as actions can be effectively represented by skeletons.}, and simultaneously extract the positional information $pos$ of each skeleton for subsequent operations, as shown in Fig.~\ref{fig:iam}-\ding{182}:
\begin{equation}
	\label{eq:Decoupling_action}
	\langle{a},\ pos\rangle = \mathrm{SE}({\rm V}),
\end{equation}
where $\mathrm{SE}$ denotes the human skeleton extraction tool\footnote{AlphaPose~\cite{li2019crowdpose} is used here; any state-of-the-art human skeleton extraction tool can be applied.}.


If action information is not removed and scene data containing actions is used directly, the action information may be considered noise, increasing the complexity of the model's processing and making the detection results unstable\footnote{The performance of the model using scene data without removed action information is shown in Table~\ref{tab:CompFW} and Table~\ref{tab:CompUN} in the ``Ours$^2$" row.}. Additionally, since the scene data contains irrelevant action information, the model may learn unrelated features, affecting its generalization ability on new data.

To prevent action information from affecting detection results, we need to remove these elements from the scene. First, using the extracted positional information $pos$, we generate an action mask $mask$ with an image segmentation tool, as shown in Fig.~\ref{fig:iam}-\ding{183}. Then, utilizing this mask with an image inpainting tool~\cite{song2022improving}, we erase the actions from the video frames, thereby obtaining clear scene data $s$, as shown in Fig.~\ref{fig:iam}-\ding{184}.

\begin{equation}
	\label{eq:Decoupling_scene}
	mask = \mathrm{ST}({\rm V}, pos),
\end{equation}
where $\mathrm{ST}$ denotes the image segmentation tool\footnote{Segment Anything Model (SAM)~\cite{kirillov2023segment} is used here; any state-of-the-art image segmentation tool can be applied.}.

\begin{equation}
	\label{eq:Decoupling_scene}
	s = \mathrm{IT}({\rm V}, mask),
\end{equation}
where $\mathrm{IT}$ denotes the image inpainting tool\footnote{Inpainting Anything Model (IAM)~\cite{yu2023inpaint} is used here; any state-of-the-art image inpainting tool can be applied.}.

Having successfully decoupled the video clips into scenes and associated human actions, we now proceed to examine the interrelationships between these elements.

\subsection{Relational Knowledge Mapper}
\label{sec:rkm}
Existing methods mostly capture and represent relationships between data through implicit associations within the learning mechanisms of the model, rather than explicitly defining and representing these relationships. For example, deep learning models learn implicit relationships between input features during training through large amounts of data and labels. These implicit relationships are reflected in the model's weights and structure but are not explicitly represented. While this is effective for some simple detection tasks, it mainly relies on automatically learned data features during training, making it difficult to fully capture and utilize complex contextual information, especially when there is insufficient training data.


As shown in Figure \ref{fig:pipeline}-\textbf{Stage 1}, we propose an explicit association method, the Relationship Knowledge Mapper (RKM) for ``Relation Interweaving". This leverages the powerful representation capabilities of knowledge graphs to explicitly integrate high-level feature, providing a deep understanding of the relationships between scenes and actions. This is crucial for improving the accuracy of anomaly detection. Additionally, this method has a flexible updating mechanism that can represent new relationships by adding new nodes and edges, thereby adapting to continuously changing data and environments.



Given the training sets, the construction of the RKM involves four processes --- clustering, combining, constructing, and updating, as shown in Fig.~\ref{fig:rkm}.

\subsubsection{Clustering}
\label{sec:Clustering}
It is unrealistic to treat all data as independent information for constructing RKM. Clustering enables us to more effectively understand and categorize complex data structures. By grouping similar scenes and actions, clustering significantly enhances the manageability and accuracy of data analysis.
For static scenes, where only the people move and the scene remains unchanged ($e.g.$, videos filmed with cameras at fixed angles),
intuitively, when we already know the number of categories\footnote{Different scene and action types categorized based on video content.} for scenes and actions, we can simply put these scenes and actions in that category and find the centers without doing clustering. In contrast, dynamic scenes feature a variable number of elements in motion, including both the scenes and the people ($e.g.$, videos captured by handheld or moving cameras), require clustering (Fig.~\ref{fig:rkm}-\ding{182}) to unify similar scenes into the same scene category, thus simplifying scene complexity and reducing scene categories. This process groups similar scenes and actions to ensure data accurately reflects the situation, while also reducing the number of scene categories, making subsequent processing more efficient.

Given any decoupled scene and human action from the dataset, we first cluster these two elements using the K-means clustering algorithm to obtain the cluster centers of the human actions and scenes from normal and abnormal videos.
We technically set the number of clustering centers of human actions within normal and abnormal videos as $\theta_{fn} $ and $\theta_{fa} $ for each clip by the distribution statistics in the datasets\footnote{Ablation studies are shown in Table~\ref{tab:ablation1}.}. The number of clustering centers of scenes is the same as the number of video scene categories.

By clustering actions and scenes, this method not only simplifies the complexity of the data but also significantly enhances processing efficiency and classification accuracy. Moreover, it strengthens the robustness and efficiency of the video analysis framework, enabling the model to perform anomaly detection more reliably when dealing with varied and complex video data.

\subsubsection{Combining}
\label{sec:Combining}
\begin{figure}[!t]
	\centering{\includegraphics[width=1\linewidth]{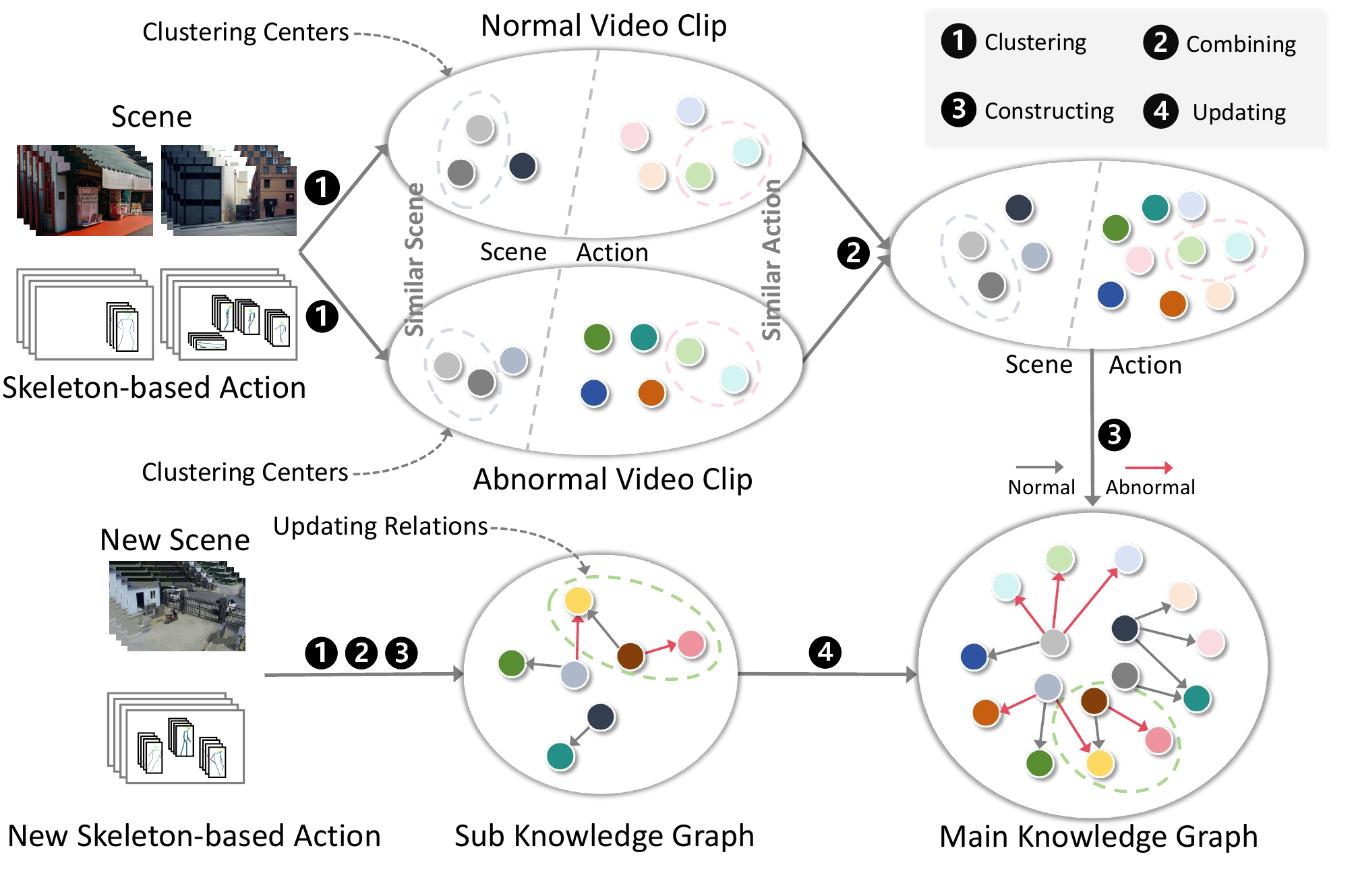}}
	\vspace{-0.8cm}
	\caption{Illustration of Relational Knowledge Mapper.}
	\label{fig:rkm}
	\vspace{-0.45cm}
\end{figure}

Since the clips of the abnormal video may contain the content of the normal actions, we combine these normal actions clustering centers with the same normal actions clustering centers in normal videos (Fig.~\ref{fig:rkm}-\ding{183}). 
This is achieved by calculating the cosine similarity (${\rm Sim}$) between these cluster centers, which is denoted by:

\begin{equation}
	\label{eq:as1}
	{\rm Sim}(\textbf{A}^{\text{fn}},\textbf{A}^{\text{fa}})=\frac{\textbf{A}^{\text{fn}}\cdot \textbf{A}^{\text{fa}}}{\big\| \textbf{A}^{\text{fn}} \big\|_2 \cdot \big\| \textbf{A}^{\text{fa}} \big\|_2 } ,
\end{equation}

where $\textbf{A}^{\text{fn}}$ and $\textbf{A}^{\text{fa}}$ denote the cluster centers of the human actions from normal videos and abnormal videos, respectively, without considering if they are normal or abnormal actions. Here, $\cdot$ represents the dot product of the vectors, and $\big\| ~ \big\|_2$ denotes the L2 norm of the vector.


Then, we combine the cluster centers of human actions from normal videos and abnormal videos --- if the cosine similarity exceeds $\rho$\footnote{The ablation study is shown in Table~\ref{tab:ablation3}-A.}, combining the two cluster centers.
These cluster centers serve as the template to guide the subsequent knowledge graph construction. Note that the cluster centers of the scenes do not need to be combined.

\subsubsection{Constructing}
\label{sec:Constructing}
In a normal video, the occurrence of an action is always considered normal, whereas in an abnormal video, the occurrence of an action may not necessarily be abnormal; it could also be normal. Thus, as shown in Fig.~\ref{fig:rkm}-\ding{184}, to construct a detailed knowledge graph, we first use normal videos' scenes and human actions and mark these relationships as ``normal''. This serves as the initial knowledge graph. 

Then, we incorporate abnormal videos' scenes and human actions into the initial knowledge graph. This is done by computing the cosine similarity between the human actions and the cluster centers in the initial knowledge graph, and based on this similarity, we assign a numerical identifier to the foreground. To achieve this process, we query the relationship between the scenes and human actions within the knowledge graph: if the relationship is ``normal'', we maintain it as is; if there is no relevant relationship, we mark it as ``abnormal". 

Let $G$ represent the initial knowledge graph consisting of a number of scene-action relationships, denoted by ($\textbf{S},\textbf{A},{\textbf{R}}$), where $\textbf{S}$ and $\textbf{A}$ are the cluster centers of the scenes and actions in normal videos, respectively, and $\textbf{R}$ is the relation between scenes and actions of normal video clips:
\begin{equation}
	\label{eq:initial}
	G=\{(\textbf{S},\textbf{A},{\textbf{R}})\},
\end{equation}
where $\textbf{R}$ is defined as ``normal'' in the initial knowledge graph. We can update the knowledge graph based on the relationships between scenes and human actions from abnormal videos:
\begin{equation}
	\label{eq:updating}
	G^{'}=\{(\textbf{S}',\textbf{A}',\textbf{R}')\},
\end{equation}
where $\textbf{S}'$ and $\textbf{A}'$ denote the cluster centers of the scenes and actions contained within both normal and abnormal video clips. $\textbf{R}'$ is the relationship between scenes and actions of normal and abnormal video clips. $\textbf{R}'$ is defined as:
\begin{equation}
	\begin{split}	
		\textbf{R}'  = \begin{cases}
			Normal,
			&\text{ $if$ } \  (\textbf{S}',\textbf{A}',\textbf{R}') \in G, \\
			Abnormal ,    &\text{ $if$ } \  (\textbf{S}',\textbf{A}',\textbf{R}') \notin G.
		\end{cases}
	\end{split}
	\label{eq:gammabeta}
\end{equation}


By querying and adjusting the relationships between scenes and human actions in the knowledge graph, these relationships can be effectively maintained or labeled as ``normal" or ``abnormal", resulting in the final knowledge graph $G^{'}$, providing support for Uncertainty Refinement (Sec.~\ref{sec:uit}).

\subsubsection{Updating} 
\label{sec:Updating}
If we want to add new video data that includes scenes and actions not previously included in the knowledge graph, we first need to construct a sub knowledge graph with the new data and then update the main knowledge graph, as illustrated in Fig.~\ref{fig:rkm}-\ding{185}. This updating process allows the knowledge graph to flexibly accommodate the inclusion of new data. This flexible knowledge graph updating mechanism provides the foundation for the system's continual learning and adaptation, enabling it to continuously adjust to evolving data and environments.

The updating process involves the dynamic generation of cluster centers based on the computation of cosine similarity between each newly added video data instance, $e.g.$, scenes and actions, and all scenes and actions cluster centers in the previously constructed knowledge graph, then, determine the maximum cosine similarity obtained, as outlined below:
\begin{equation}
	\label{eq:as1}
	{\rm max}_{sim}^{a} = {\rm Max}\big( {\textstyle \bigcup_{i}^{n}} {\rm Sim}(\textbf{A}^{\text{new}}_{i},\textbf{A}')\big),
\end{equation}
\begin{equation}
	\label{eq:as2}
	{\rm max}_{sim}^{s} = {\rm Max}\big( {\textstyle \bigcup_{i}^{n}} {\rm Sim}(\textbf{S}^{\text{new}}_{i},\textbf{S}')\big),
\end{equation}
where $\textbf{A}^{\text{new}}_{i}$ and $\textbf{S}^{\text{new}}_{i}$ are the newly added $i$-th action and scene. ${\rm Sim}$ denotes the cosine similarity. ${\rm Max}$ is the maximization operation to obtain the maximal value of cosine similarity of actions (${\rm max}_{sim}^{a}$) and scenes (${\rm max}_{sim}^{s}$). ${\textstyle \bigcup_{i}^{n}}$ is the union of the values of cosine similarity. $n$ means the total number of newly-added actions or scenes.


Based on the calculation results of the maximum cosine similarity, we add the newly added  $i$-th action and scene as new cluster centers into $\textbf{A}'$ and $\textbf{S}'$, denoted as $add$.
\begin{equation}
	\begin{split}	
		\begin{cases}
			\textbf{A}^{\text{new}}_{i}\overset{\rm add}{\rightarrow} \textbf{A}',
			&\text{ $if$ }\ {\rm max}_{sim}^{a} \le \mu_a, \\
			\textbf{S}^{\text{new}}_{i}\overset{\rm add}{\rightarrow} \textbf{S}' ,    
			&\text{ $if$ }\ {\rm max}_{sim}^{s} \le \mu_s,
		\end{cases}
	\end{split}
	\label{eq:gammabeta}
\end{equation}
where $\mu_a$ and $\mu_s$ are thresholds to determine the $add$ operation. The ablation study of these two thresholds can be seen in Table~\ref{tab:ablation4}.
It's important to note that this process makes no distinction between normal and abnormal video clips.

Then, when the maximal value of cosine similarity of actions (${\rm max}_{sim}^{a}$) and scenes (${\rm max}_{sim}^{s}$) are greater than $\mu$, we combine the newly-added $i$-th action and scene into $\textbf{S}'$ and $\textbf{A}'$, denoted by $combine$, with existing cluster centers in the constructed knowledge graph:
\begin{equation}
	\begin{split}	
		\begin{cases}
			\textbf{A}^{\text{new}}_{i}\overset{\rm combine}{\rightarrow} \textbf{A}',
			&\text{ $if$ }\ {\rm max}_{sim}^{a} > \mu_a, \\
			\textbf{S}^{\text{new}}_{i}\overset{\rm combine}{\rightarrow} \textbf{S}',    
			&\text{ $if$ }\ {\rm max}_{sim}^{s} > \mu_s.
		\end{cases}
	\end{split}
	\label{eq:gammabeta}
\end{equation}

Moreover, directly updating the main knowledge graph with all the relationships from the sub knowledge graph might lead to a decline or even failure in the model's detection capability, as there could be extreme or incorrect relationships in the sub knowledge graph. Therefore, we need to filter the relationships in the sub knowledge graph by calculating the cosine similarity between the nodes of the sub relationships and the nodes of the main relationships. If the sub relationship with the highest cosine similarity matches the main relationship, we proceed with the update; otherwise, we do not update the relationship. This ensures the safe updating of the main knowledge graph. It is important to note that all nodes in both the sub knowledge graph and the main knowledge graph come from $\textbf{S}'$ and $\textbf{A}'$.

In this way, we complete the construction of the detailed knowledge graph for ``Relation Interweaving'' to obtain scene-action relations. Next, we will detail how to use ``Feature Interweaving'' to obtain initial anomaly scores.

\subsection{Scene-Action Integrator}
\label{sec:sad}

As shown in Fig.~\ref{fig:pipeline}-\textbf{Stage 1} (Step2), to enhance video anomaly detection involving human subjects, we introduce the Scene-Action Integrator (SAI) for ``Feature Interweaving". This innovative approach scrutinizes individual motion and posture and comprehensively interprets the environmental context. SAI represents a multifaceted strategy that effectively bridges the gap between human actions and their surroundings, leveraging a deeper understanding of physical movements and environmental semantics.

To implement the SAI, we use the decoupled scenes ($\text{sc}$) and the isolated human action ($\text{sk}$) from the video clips. Using skeleton features, we encode the scenes with a feature encoder ($\mathcal{E}$) and capture semantic relationships with a Graph Convolution Network (GCN) operation ($\mathcal{G}$). To understand temporal dynamics, we employ a Long Short-Term Memory (LSTM) network ($\mathcal{LM}$). Position embeddings ($\mathcal{PE}$) record the position of the actions within previous scenes, ensuring coherent integration and reasonable action arrangement when fusing with another action. By concatenating the features through the operation ($\mathcal{C}$) to obtain the fused features $f_{concat}$, and passing them through the fully-connected layer ($\mathcal{FC}$), we obtain the final anomaly scores (AS). This approach combines skeleton-based representations, semantic relationships, temporal dynamics, and positional information to generate accurate anomaly scores.
The whole processing is denoted by:

\begin{equation}\small
	\begin{aligned}
		&{\rm AS} = \mathcal{FC} (\underset{\Uparrow}{\underline{f_\emph{concat}}}).\\[-0.15cm]
		&\hspace{-0.315cm}\overbrace{\rm{{\mathcal{C}(\mathcal{E}(\rm sc), \mathcal{LM}(\mathcal{G}(\rm sk)),\mathcal{PE}(\rm sk))}}}
	\end{aligned}
\end{equation}

%

In training SAI, we employ the Multiple Instance Learning approach. As illustrated in the upper right of Fig.~\ref{fig:pipeline}, consider a typical video composed of multiple clips. Each clip is assigned an anomaly score. To determine the anomaly score for the entire video\footnote{We compile N clips from each normal video into a normal bag, while N clips from an abnormal video are grouped into an abnormal bag. Each clip contains 24 frames. The ablation study is shown in Table~\ref{tab:ablation3}-B.}. We select and average the highest $K$ anomaly scores from these clips. This method is applied consistently to both normal and abnormal videos.

This procedure effectively increases the distinction between normal and abnormal videos by amplifying the difference in their respective anomaly scores. This approach is instrumental in enhancing the model's ability to differentiate between normal and abnormal content in video data. 

\subsection{Uncertainty Refinement}
\label{sec:uit}
We propose Uncertainty Refinement(UR) to train our DecoAD in an iterative training way in \textbf{Stage 2}\footnote{The ablation study is shown in Table~\ref{tab:ablation3}-C.} (Step3). To achieve this goal, we set hyperparameters $\beta_{1}$ and $\beta_{2}$ as thresholds\footnote{The ablation study is shown in Table~\ref{tab:ablation2}.} and construct three pools, $i.e.$, ``normal pool'', ``abnormal pool'' and ``pending pool''. Initially, the ``normal pool'' is constructed by normal video clips. For abnormal video clips, we first combine all scenes (including their positional information) with the human actions and feed them into the models of the \textbf{Stage 1}. In the first iteration (\textbf{Stage 2}), the abnormal video clips are further put into these three pools based on the anomaly scores and the relationships in the knowledge graph:

1) Video clips with anomaly scores below $\beta_{1}$ and marked as ``normal" in the knowledge graph $G^{'}$ are placed in the ``normal pool", as normal training datas;

2) Video clips with anomaly scores above $\beta_{2}$ and marked as ``abnormal" in the knowledge graph $G^{'}$ are placed in the ``abnormal pool", as abnormal training datas;

3) Video clips that do not meet the above two conditions are placed in the ``pending pool", which is used for UR iteration training.
Then, we use the data from the ``pending pool" for further iterative training of the model. 

\begin{figure}[!t]
	\centering{\includegraphics[width=1\linewidth]{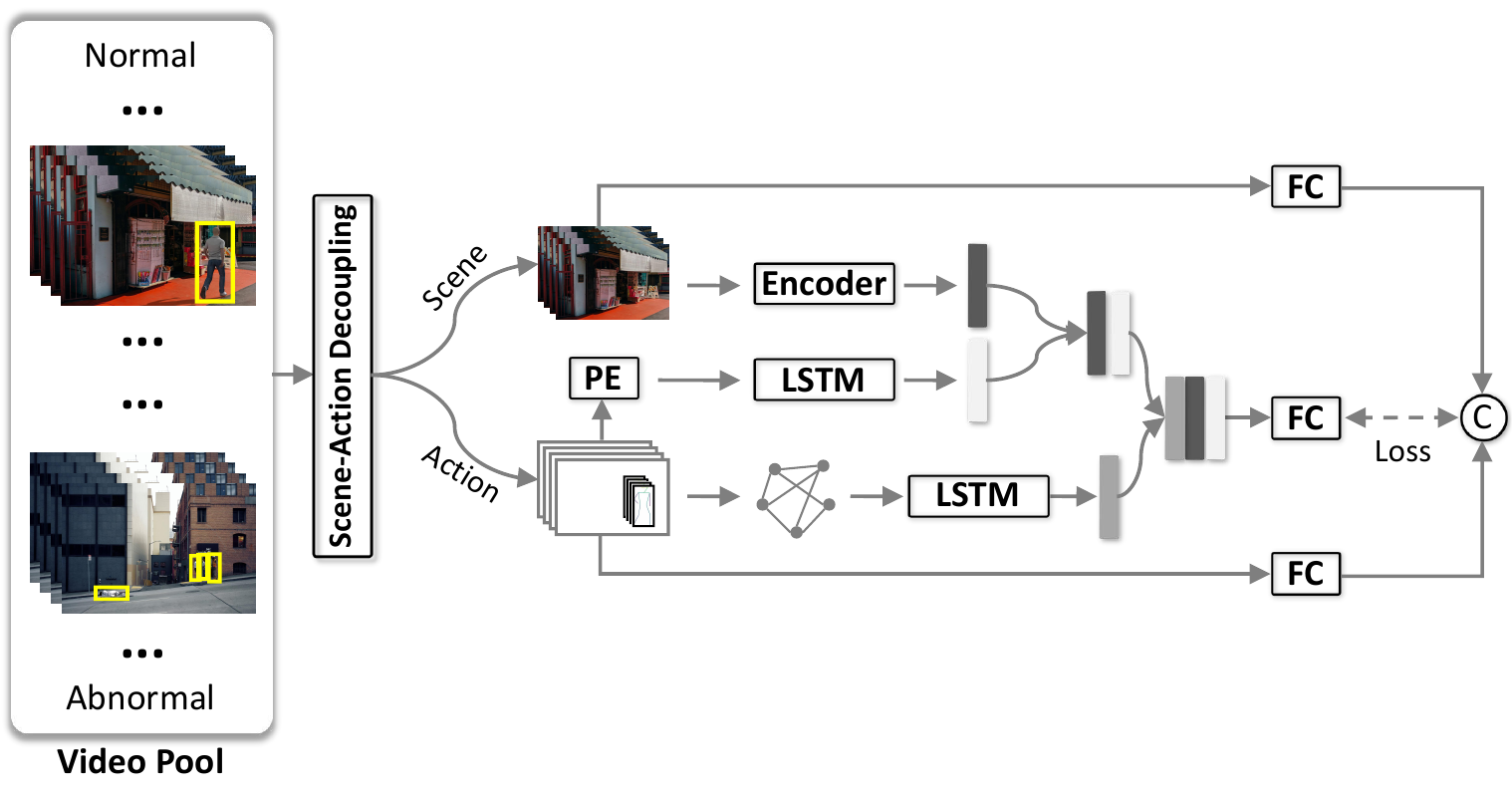}}
	\vspace{-0.5cm}
	\caption{Pipeline of unsupervised training, which is based on the traditional auto-encoder, using improved Scene-Action Integrator (Sec.~\ref{sec:sad}) as the backbone.}
	\label{fig:pipelineUN}
	\vspace{-0.45cm}
\end{figure}

\subsection{Training Methodology}
The method mentioned above is trained under fully-supervised and weakly-supervised conditions. To increase the generalization, our method can also be trained in an unsupervised learning manner. In the unsupervised learning environment, where the training phase involves only normal videos, which does not meet the requirements of Multiple Instance Learning, we instead employ a traditional auto-encoder~\cite{hinton2006reducing} to tackle this challenge. As shown in Fig.~\ref{fig:pipelineUN}, we utilize the original model (SAI) as the encoder and construct a corresponding decoder within this framework. By comparing the combined features of the input videos with the reconstructed video features, we can determine the presence of anomalies.

Inspired by the knowledge graph, we adopt a similar strategy of recombining all scenes and human actions. This is done to maximize the auto-encoder's grasp and learning of the features within normal video clips, thus enhancing its capability for detecting abnormal situations.

Note that the main differences between unsupervised and fully/weakly-supervised training methodology are two manifolds --- 1) The Scene-Action Integrator (Sec.~\ref{sec:sad}) in \textbf{Stage 1} (Step2), where in unsupervised training, it changes to an auto-encoder; 2) The Relational Knowledge Mapper in \textbf{Stage 1} (Step1) and UR in \textbf{Stage 2} (Step3) are discarded from fully/weakly-supervised training.

\subsection{Training Loss}
\textbf{Fully-supervised and Weakly-supervised Training}.
In \textbf{Stage 1} of both fully-supervised and weakly-supervised training, we calculate the Multiple Instance Learning Loss~\cite{sun2016multiple}, denoted as $\mathcal{L}_{mil}$, by comparing the anomaly scores of abnormal videos with those of normal videos. The overall process can be formulated as follows:
\begin{equation}
	\begin{aligned}
		\mathcal{L}_{mil} =\alpha_1 \times \mathcal{L}_{rank} + \alpha_2 \times \mathcal{L}_{focal},
	\end{aligned}	
\end{equation}
where $\alpha_1$ and $\alpha_2$ are learnable weight parameters. $\mathcal{L}_{rank}$ is the Ranking Loss~\cite{schroff2015facenet}. $\mathcal{L}_{focal}$ is the Focal Loss~\cite{lin2017focal} incorporating with BCE Loss.

%

In \textbf{Stage 2}, to train our DecoAD iteratively under fully/weakly-supervised conditions, we employ the Binary Cross-Entropy loss ($\mathcal{L}_{bce}$) to increase the distance between the ``normal pool" and the ``abnormal pool". The total loss ($\mathcal{L}_{total}$) in this stage is formulated as:
\begin{equation}
	\label{eq:loss_total_fw}
	\mathcal{L}_{total} = \lambda_1 \times \mathcal{L}_{mil}+\lambda_2 \times \mathcal{L}_{bce}.
\end{equation}
where $\lambda_1$ and $\lambda_2$ are learnable weight parameters.

\textbf{Unsupervised training}.
For unsupervised training, we have excluded the Relational Knowledge Mapper and the Uncertainty Refinement and modified the Scene-Action Integrator to an autoencoder (Fig.~\ref{fig:pipelineUN}).
The total loss ($\mathcal{L}_{total}$) for unsupervised training are consisting of reconstruction loss ($\mathcal{L}_{rec}$) and regularization term ($\mathcal{L}_{reg}$) is formulated as:
\begin{equation}
	\label{eq:loss_total_un}
	\mathcal{L}_{total} = \lambda_1 \times \mathcal{L}_{rec}+\lambda_2 \times \mathcal{L}_{reg}.
\end{equation}
where $\lambda_1$ and $\lambda_2$ are learnable weight parameters. The regularization term $\mathcal{L}_{reg}$ is calculated using L2 regularization to prevent overfitting by penalizing large weights in the model.

\section{Experiments}
\subsection{Datasets}
We evaluate our method on three datasets, namely NWPU Campus~\cite{cao2023new}, UBnormal~\cite{acsintoae2022ubnormal}, and HR-ShanghaiTech~\cite{liu2018future}. According to the characteristics of each dataset, we employ UBnormal for fully/weakly-supervised training, NWPU Campus for weakly-supervised training, and NWPU Campus, UBnormal, and HR-ShanghaiTech for unsupervised training.

The NWPU Campus dataset includes 43 different scenes and 28 types of abnormal events, pioneering the study of scene-dependent anomalies. However, its training set only contains normal video data, which does not meet the requirements for weakly supervised video anomaly detection. Therefore, we reconfigured the training and test sets to accommodate weakly supervised models, but we still used the original dataset for unsupervised training. The UBnormal dataset comprises 29 scenes and 22 types of abnormal events, with detailed annotations that make it highly valuable for advanced anomaly detection research. HR-ShanghaiTech, a subset of the ShanghaiTech Campus dataset, focuses on human-related scenes, encompassing 13 scenes and 11 types of abnormal events.

\subsection{Evaluation Metrics}

In the field of video anomaly detection, the commonly used performance evaluation metric is the area under the Receiver Operating Characteristic curve (AUC), which intuitively reflects the performance of detection methods. However, due to the imbalance in anomaly detection tasks, AUC may exaggerate performance. Therefore, we introduce the area under the Precision-Recall curve (AP) as a supplementary metric. A higher AP value indicates a stronger ability of the model to detect abnormal events.

%
%
%

\subsection{Implementation Details}
Our work is implemented in PyTorch and experimented on NVIDIA RTX 4090 GPU. We employ the AlphaPose~\cite{li2019crowdpose} and YOLOX~\cite{ge2021yolox} detectors to independently detect the human skeleton in each video frame. The network is optimized using the Adam optimizer ($\beta _{1}=0.9$, $\beta _{2}=0.999$) with an initial learning rate of $1\times 10^{-4}$ for all model training, which decreases by multiplying 0.1 for every 10 epochs. Our method utilizes a batch size of 256, and the training process runs for a total of 120 epochs, only costing 2.2 hours. Additionally, the size of our supervised model has been optimized to 1 Mb, while the unsupervised model size has been optimized to 12.3 Mb, with the frames per second (FPS) remaining around 24.

\begin{table}[!t]
	\centering
	\caption{Quantitative evaluation of major components used in our approach in terms of the AUC (\%) performance on the UBnormal (UB) dataset. The best results are marked in \textbf{bold}.}
	\vspace{-0.2cm}
	\begin{tabular}{c}
		\hspace{-0.1cm}
		\includegraphics[width=0.95\linewidth]{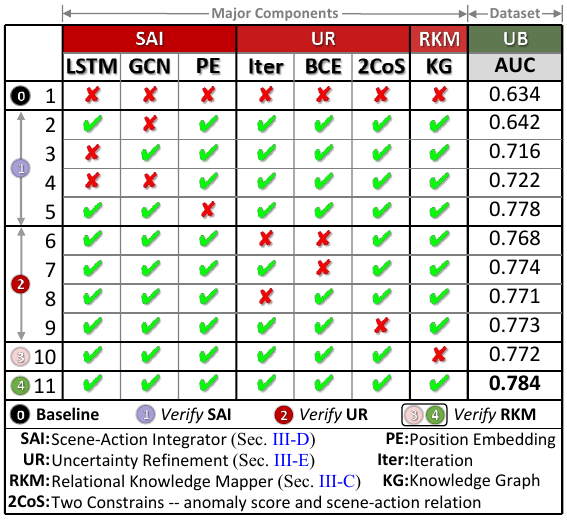}
	\end{tabular}
	\label{tab:component}
	\vspace{-0.5cm}
\end{table}

\subsection{Component Evaluation}
We conducted a comprehensive evaluation of our method's components, as shown in Table~\ref{tab:component}. To ensure successful code execution, we replaced the key components requiring verification with simpler operations. For example, we substituted the proposed components with a basic ResNet model~\cite{he2016deep} consisting of two fully connected layers. This served as our baseline, and the qualitative results are shown in line 1.

Lines 2-5 demonstrate the effectiveness of the Scene-Action Integrator (Sec.~\ref{sec:sad}) in achieving ``Feature Interweaving" between scenes and associated human actions. Comparing line 4 (our method) to line 11, where we removed LSTM and GCN, we observed a decrease in the area under the curve (AUC) from 78.4\% to 72.2\%. Additionally, we observed that line 3 (GCN) outperformed line 2 (LSTM), with AUC values of 64.2\% and 71.6\%, respectively, indicating that GCN is better at modeling action relationships, which is crucial for understanding human actions. These results underscore the importance of the Scene-Action Integrator in capturing the relationship between scenes and human actions, and highlight the effectiveness of GCN in this task.

Lines 6-9 provide evidence of the effectiveness of Uncertainty Refinement (Sec.~\ref{sec:uit}). By comparing line 7 to line 8, we deduced that the iterative training process of the ``pending" pool is more effective than using binary cross-entropy (BCE) loss for the ``normal" pool and sub-``abnormal" pool, as indicated by the higher AUC. Moreover, removing the two constraints on anomaly score and scene-action relation (line 9) resulted in decreased AUC performance.

Comparing line 10 to line 11, our method incorporating the Relational Knowledge Mapper (Sec.~\ref{sec:rkm}, line 11) outperforms the method without it (line 10). This is because the Relational Knowledge Mapper enables a comprehensive understanding of the intricate interplay between different scenes and human actions by leveraging a detailed knowledge graph.


\subsection{Performance Comparison}
To demonstrate the effectiveness of our approach, we conducted a comprehensive comparison with state-of-the-art methods using three different training methodologies: fully-supervised, weakly-supervised, and unsupervised training.

For fully/weakly-supervised training, we selected the DeepMIL~\cite{sultani2018real}, ST-GCN~\cite{yan2018spatial}, Shift-GCN~\cite{cheng2020skeleton}, RTFM~\cite{tian2021weakly}, MGFN~\cite{chen2023mgfn}, BN-WVAD~\cite{zhou2023batchnorm}, STG-NF~\cite{hirschorn2023normalizing}, and RTFM-BERT~\cite{tan2024overlooked}. For unsupervised training, we evaluated the GEPC~\cite{markovitz2020graph}, MPN~\cite{lv2021learning}, LGN-Net~\cite{zhao2022lgn}, MoCoDAD~\cite{flaborea2023multimodal}, STG-NF~\cite{hirschorn2023normalizing}, CampusVAD~\cite{cao2023new}, TrajREC~\cite{stergiou2024holistic}, and GiCiSAD~\cite{karami2024graph} methods. 
The results we compared were obtained either from the source code or reported results provided by the respective authors.
The ``Ours$^{1}$" is our method which does not consider scene information, meaning that the model only utilizes skeleton information for video anomaly detection and cannot perform Relational Knowledge Mapper (RKM) construction or Uncertainty Refinement (UR). The ``Ours$^2$" is our method, but it uses scene data for training without removed action information, as detailed in Sec.~\ref{sec:decoup}. The ``Ours*" comprehensively considers all information (skeleton, scene, and location).

\begin{table}[!t]
	\centering
	\caption{Quantitative performance comparison with other state-of-the-art methods on the NWPU Campus (denoted by NWPUC, used for weakly-supervised training) and UBnormal (denoted by UB, used for fully/weakly-supervised training) datasets, regarding frame-level AUC and AP metrics in fully/weakly-supervised training (denoted by ``weakly'' and ``fully''); Red color represents the best, and green color represents the second best; FPS stands for frames per second. }
	\vspace{-0.2cm}
	\begin{tabular}{c}
		\hspace{-0.1cm}
		\includegraphics[width=0.95\linewidth]{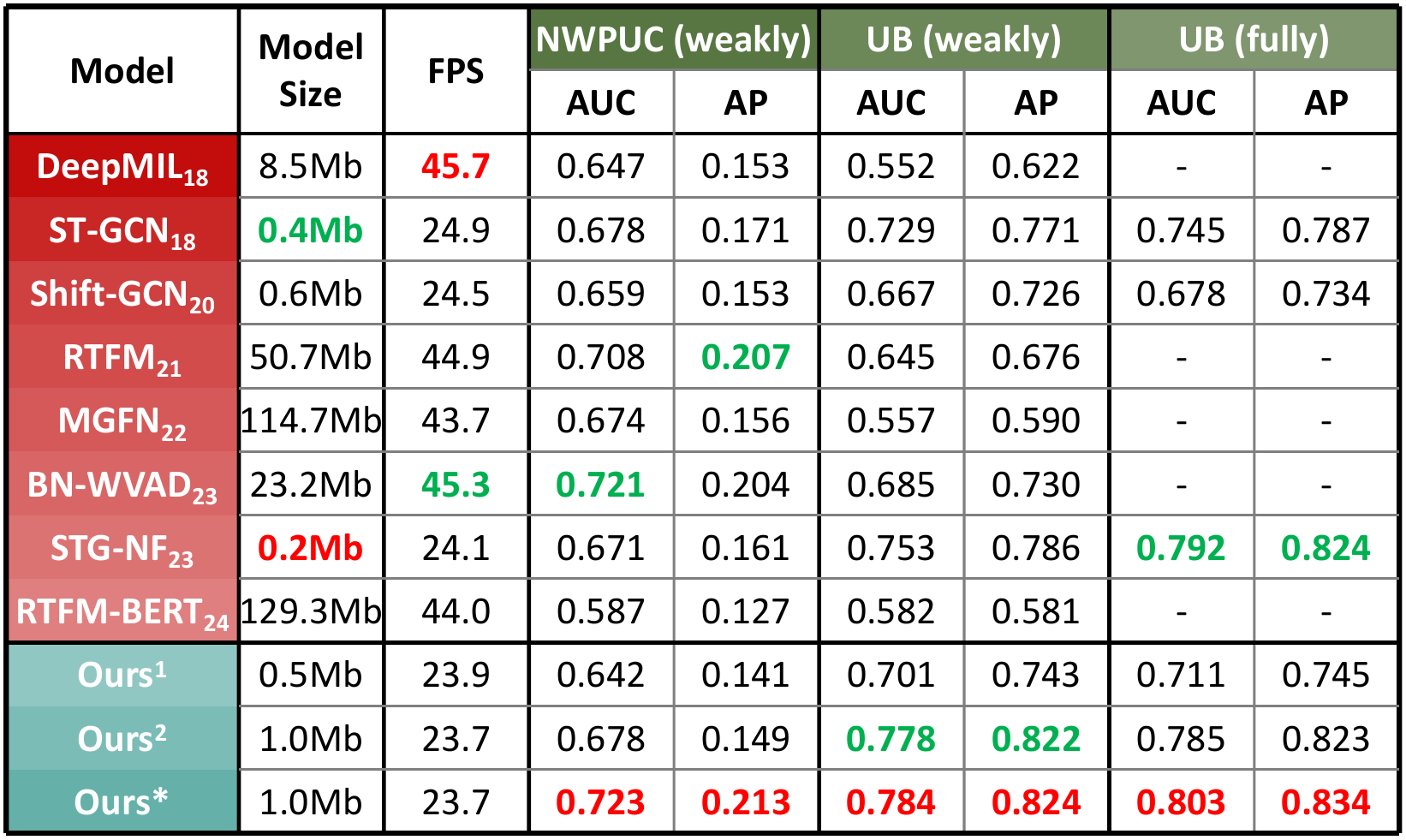}
	\end{tabular}
	\label{tab:CompFW}
	\vspace{-0.25cm}
\end{table}

\begin{table}[!t]
	\centering
	\caption{Quantitative performance comparison with other state-of-the-art methods on the NWPU Campus (NWPUC), UBnormal (UB), and HR-ShanghaiTech (HR-STC) datasets, regarding frame-level AUC and AP metrics in unsupervised training (denoted by ``un''); Red color represents the best, and green color represents the second best; FPS stands for frames per second.}
	\vspace{-0.2cm}
	\begin{tabular}{c}
		\hspace{-0.1cm}
		\includegraphics[width=0.95\linewidth]{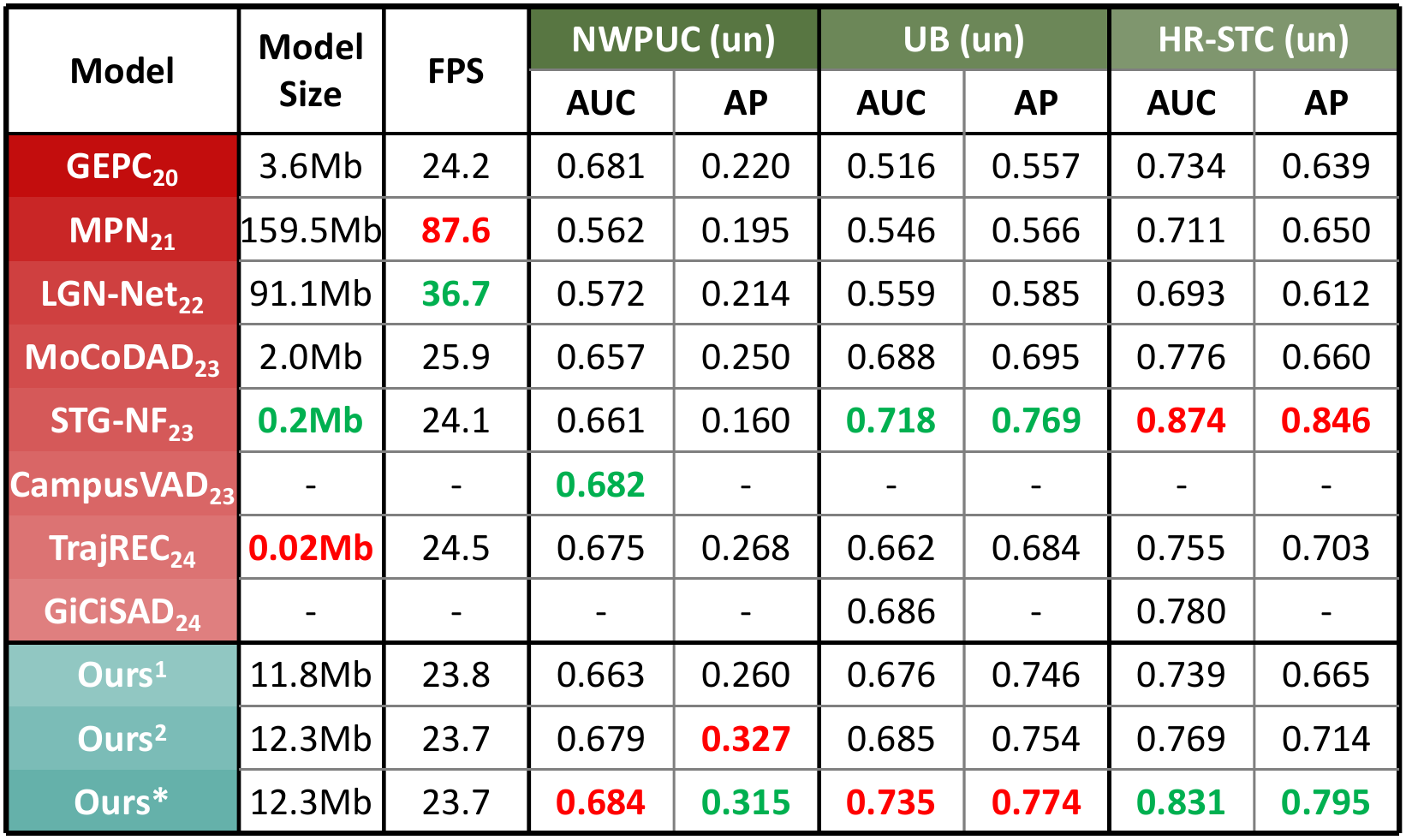}
	\end{tabular}
	\label{tab:CompUN}
	\vspace{-0.55cm}
\end{table}



\subsubsection{Quantitative Comparisons with Fully/Weakly-supervised Training Methods}
The quantitative comparison results with fully/weakly-supervised training methods are shown in Table~\ref{tab:CompFW}. We found that ``Ours$^{1}$" shows inferior performance compared to existing action-based methods such as STG-NF. STG-NF overlooks scene information, operating directly on the distribution of data and providing a more direct probabilistic interpretation, making it more sensitive to the detection of abnormal behaviors. Our proposed method ``Ours*" outperforms all previous state-of-the-art approaches in fully/weakly-supervised training settings. Specifically, ``Ours*" achieves an improvement of 0.2\% and 3.1\% in AUC values, and 0.6\% and 3.8\% in AP values over the best existing weakly-supervised methods on NWPU Campus and UBnormal, respectively. Moreover, it achieves an improvement of 1.1\% in AUC value and 1.0\% in AP value over the best existing fully-supervised method on UBnormal. These results demonstrate the effectiveness of our proposed method, which leverages the ``Scene-Action Interweaving" approach to combine and analyze elements from different scenes and human actions in videos for enhanced anomaly detection.

\begin{figure*}[!t]
	\centering{\includegraphics[width=1\linewidth]{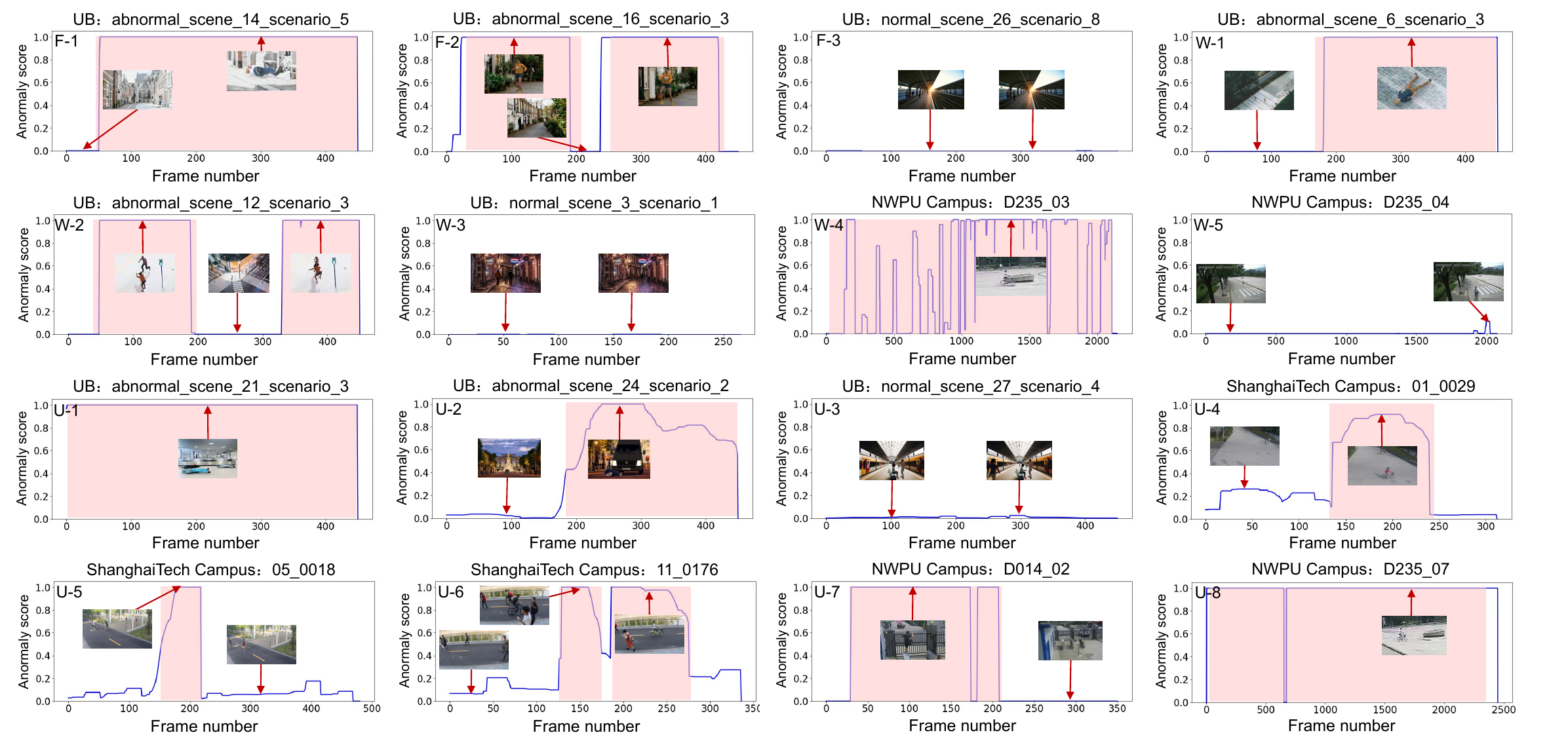}}
	\caption{The qualitative results of our method on testing videos. Colored windows indicate the true abnormal regions. ``F'': Fully-supervised; ``W'': Weakly-supervised; ``U'': Unsupervised.}
	\label{fig:quality}
	\vspace{-0.45cm}
\end{figure*}

\begin{table}[!t]
	\centering
	\caption{Ablation study on different clustering center numbers (Sec. \ref{sec:Clustering}) on the UBnormal dataset. ``$\theta_{fn} $" and ``$\theta_{fa}$": clustering center number of human actions within normal and abnormal video clips.}
	\vspace{-0.2cm}
	\begin{tabular}{c}
		\hspace{-0.3cm}
		\includegraphics[width=0.85\linewidth]{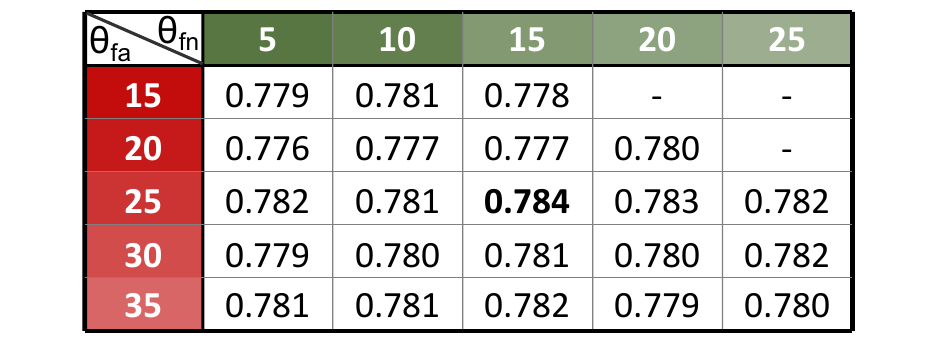}
	\end{tabular}
	\label{tab:ablation1}
	\vspace{-0.25cm}
\end{table}

\begin{table}[!t]
	\centering
	\caption{Ablation study on different thresholds for constructing three pools (Sec.~\ref{sec:uit}) on the UBnormal dataset. ``$\beta_{1}$" and ``$\beta_{2}$": different thresholds used to divide the three pools.}
	\vspace{-0.2cm}
	\begin{tabular}{c}
		\hspace{-0.3cm}
		\includegraphics[width=0.85\linewidth]{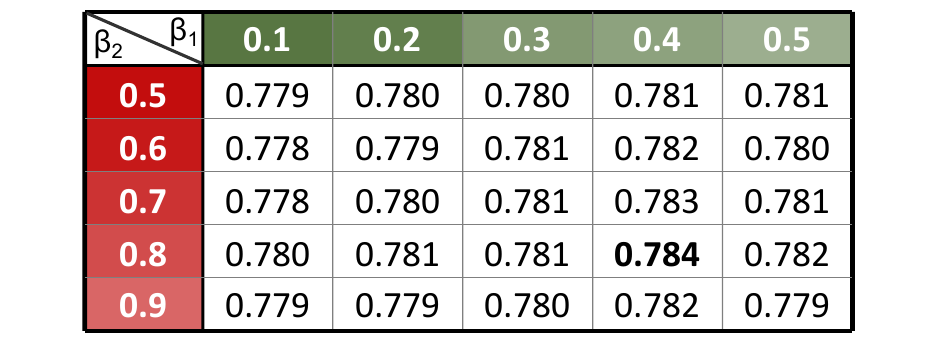}
	\end{tabular}
	\label{tab:ablation2}
	\vspace{-0.55cm}
\end{table}

\subsubsection{Quantitative Comparisons with Unsupervised Training Methods}
The quantitative comparison results with unsupervised training methods are shown in Table~\ref{tab:CompUN}. We found that the ``Ours$^{1}$" method performs worse than existing action-based methods such as TrajREC. The TrajREC method overlooks scene information and directly uses skeleton data, utilizing a self-supervised learning approach to enhance reinforcement learning effectiveness through positive and negative sample pairs. This strategy improves the model's ability to distinguish between normal and abnormal trajectory behaviors. 
Meanwhile, we found that ``Ours$^2$" performs worse than ``Ours*" because the use of scene data containing action information interfered with the model's training, thereby affecting its performance. Our ``Ours*" method also surpasses all previous state-of-the-art unsupervised training methods in NWPU Campus and UBnormal. ``Ours*" achieves improvements of 0.2\% and 1.7\% in AUC values, and 4.7\% and 0.5\% in AP values over the best existing unsupervised method, MoCoDAD, on the NWPU Campus and UBnormal datasets, respectively. Additionally, Our method achieves suboptimal results on the HR-ShanghaiTech dataset. Although some methods have smaller model sizes and higher FPS values, their video anomaly detection capabilities are not excellent. Our method (both supervised and unsupervised), after balancing model size, FPS, and video anomaly detection capability, achieves the best performance.

\subsubsection{Qualitative Results}
Fig.~\ref{fig:quality} demonstrates the superior results of our method (fully/weakly-supervised and unsupervised) in context-related situations. Our approach successfully and promptly detects these abnormal events by generating high anomaly scores for abnormal frames. F-3, W-3, W-5, and U-3 are four normal videos, for which our method generates low anomaly scores throughout the entire video (close to 0). It is worth mentioning that W-4 depicts a person riding a bicycle in a square, while W-5 shows a person riding a bicycle on a bike lane. The former is an abnormal event, while the latter is a normal event. Our model successfully identifies and detects this abnormal event in the scene without any false alarms, thanks to the concept of ``Scene-Action Interweaving".

\subsection{Ablation Study}
\subsubsection{Choices of the Number of Cluster Centers}

Since the clustering operation in the Rational Knowledge Mapper (see Sec. \ref{sec:Clustering}) is to unify similar scenes into the same scene category, thus simplifying scene complexity and reducing scene categories, the number of cluster centers is not ideal if it's too large or too small. Thus, we conducted an ablation study on the UBnormal dataset to determine the proper number of cluster centers. As shown in Table \ref{tab:ablation1}, when the number of cluster centers is too small, it fails to distinguish effectively between very similar scenes or actions, reducing the efficacy of the model. 
Conversely, when the number of cluster centers is too big, although a more refined data segmentation is possible, it may lead to model overfitting, where the features learned are too specific and fail to generalize to new data.
Thus, we set the number of cluster centers for human actions in normal and abnormal video segments to 15 and 25 respectively to achieve sufficient coverage and distinction.

\subsubsection{Choices of $\beta_{1}$ and $\beta_{2}$ in Constructing Three Pools}
Additionally, we conducted further experiments on the UBnormal dataset to explore the impact of different thresholds on the classification of the ``pending pool" (see Sec.~\ref{sec:uit}).
Proper threshold settings help the model generalize better to new and unseen data. Setting the thresholds too high or too low could lead to inappropriate sensitivity of the model to the data, thereby affecting its performance in practical applications. As shown in Table \ref{tab:ablation2}, when $\beta_{1}$ was set too low, normal video clip data might incorrectly classify as abnormal; conversely, if $\beta_{1}$ is too high, abnormal data might be wrongly classified as normal, thus reducing the overall performance of DecoAD. 
Our DecoAD achieved its best performance when $\beta_{1}$ and $\beta_{2}$ were set to 0.4 and 0.8, respectively. This is primarily because these thresholds effectively differentiated between normal and abnormal data within the ``pending pool". 

\subsubsection{Choices of Different Cosine Similarity Threshold $\rho$ in Combining Two Cluster Centers}
We conducted another ablation study on the UBnormal dataset to examine the effect of different cosine similarity thresholds on combining two cluster centers~(Sec.~\ref{sec:Combining}). As shown in Table \ref{tab:ablation3}-A, the results indicated that when $\rho$ was 0.95, the clustering result was closest to the true number of categories. Therefore, we set it as the cosine similarity threshold for DecoAD. Moreover, DecoAD achieved the best results on this basis, possibly because this threshold allowed the merged cluster centers to align more closely with the distribution of human actions in the actual dataset.

\subsubsection{Choices of Different Segment Lengths of Video Clips}
We notice that the frame rates of the datasets we compared vary. For instance, the UBnormal dataset is at 30 fps, while the HR-ShanghaiTech and NWPU Campus datasets are at 24 fps. To evaluate the effectiveness of different segment lengths of video clips (see Sec.~\ref{sec:sad}), we conducted extensive experiments.
Segment length is a critical factor in determining the time window observed by the model when making decisions. If the segment length is too short, it may not capture enough behavior sequences, making it difficult to accurately understand the context of the behavior. If it's too long, it might introduce redundant information, reducing processing efficiency and complicating the extraction of key features. The right segment length helps maintain the continuity of behavior and avoids interference from irrelevant actions or background activities, enhancing the model's recognition capabilities. 
As shown in Table \ref{tab:ablation3}-B, we found that setting the segment length to 24 frames offers the best performance, while settings of 12 or 30 frames led to significant performance declines. A 24-frame length strikes the perfect balance between the comprehensiveness of data and the complexity of processing, allowing the DecoAD model to achieve optimal performance on these specific datasets.

\begin{table}[!t]
	\centering
	\caption{Ablation study on different cosine similarity thresholds for fusing two clustering centers (\textbf{A}) (Sec.~\ref{sec:Combining}), different segment lengths (\textbf{B}) (Sec.~\ref{sec:sad}), and different iteration times (\textbf{C}) (Sec.~\ref{sec:uit}). ``$\rho$'': cosine similarity threshold; ``f'': video clip frame numbers; ``t'': iteration times of the stages; NWPUC represents the NWPU Campus dataset, UB represents the UBnormal dataset, and HR-STC represents the HR-ShanghaiTech dataset.}
	\vspace{-0.2cm}
	\begin{tabular}{c}
		\hspace{-0.3cm}
		\includegraphics[width=0.85\linewidth]{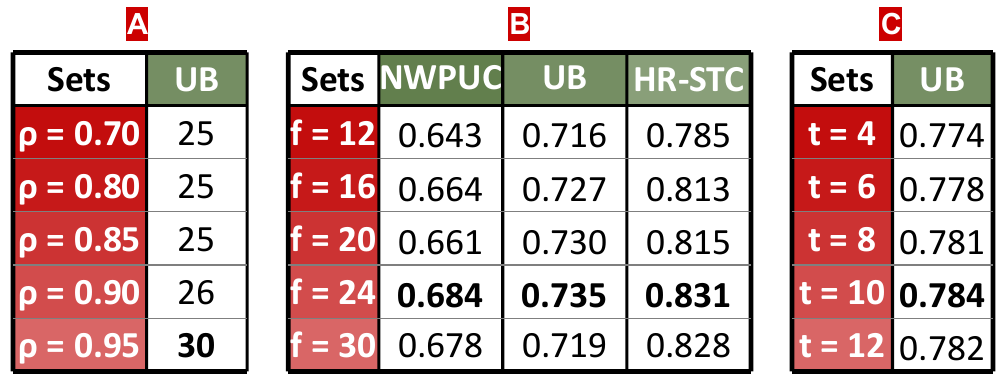}
	\end{tabular}
	\label{tab:ablation3}
	\vspace{-0.25cm}
\end{table}

\begin{table}[!t]
	\centering
	\caption{Ablation study on the updating cosine similarity thresholds $\mu_a$ for actions and $\mu_s$ for scenes (Sec.~\ref{sec:rkm}); UB represents the UBnormal dataset.}
	\vspace{-0.2cm}
	\begin{tabular}{c}
		\hspace{-0.3cm}
		\includegraphics[width=0.85\linewidth]{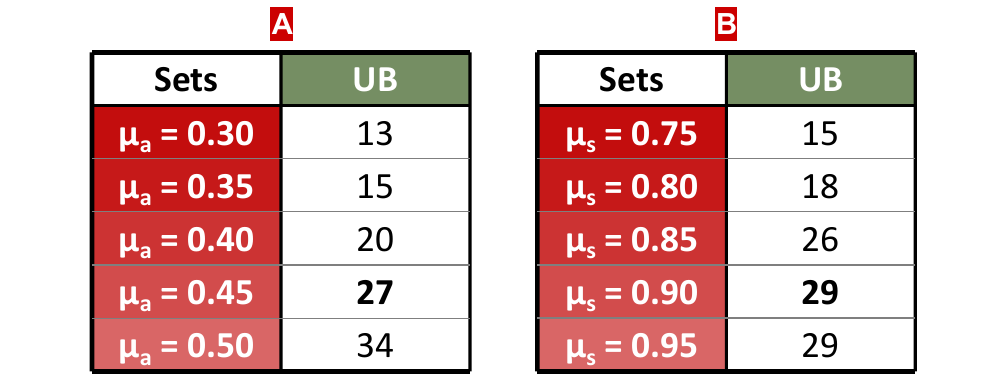}
	\end{tabular}
	\label{tab:ablation4}
	\vspace{-0.59cm}
\end{table}

\subsubsection{Effectiveness of the Number of Iterations}
We also conducted comprehensive experiments to assess the effectiveness of iteration numbers in the uncertainty refinement process (see Sec.~\ref{sec:uit}). As shown in Table \ref{tab:ablation3}-C, performance improved with an increase in iterations. However, after reaching ten iterations, the performance began to stabilize. This phenomenon could be due to insufficient data in the ``pending pool", making it difficult to further effectively expand the ``normal pool" and ``abnormal pool", and the model may have already converged to its potential optimal solution.

\subsubsection{Effectiveness of the Updating Thresholds}
We further conducted an ablation study on the updating thresholds $\mu_a$ (for actions) and $\mu_s$ (for scenes) (see Sec.~\ref{sec:rkm}). To determine the updating thresholds, we carried out ablation experiments on the same dataset. As shown in Table \ref{tab:ablation4}, we found that when $\mu_a$ was set to 0.45, the number of action clusters was closest to the actual number of action categories. Similarly, when $\mu_s$ was set to 0.90, the number of scene clusters was closest to the actual number of scene categories, indicating that the updating effect was optimal at these thresholds.

\subsection{In-depth Discussion of the Poor AP Performance}
\label{sec:lim}
\begin{figure}[!t]
	\centering{\includegraphics[width=1\linewidth]{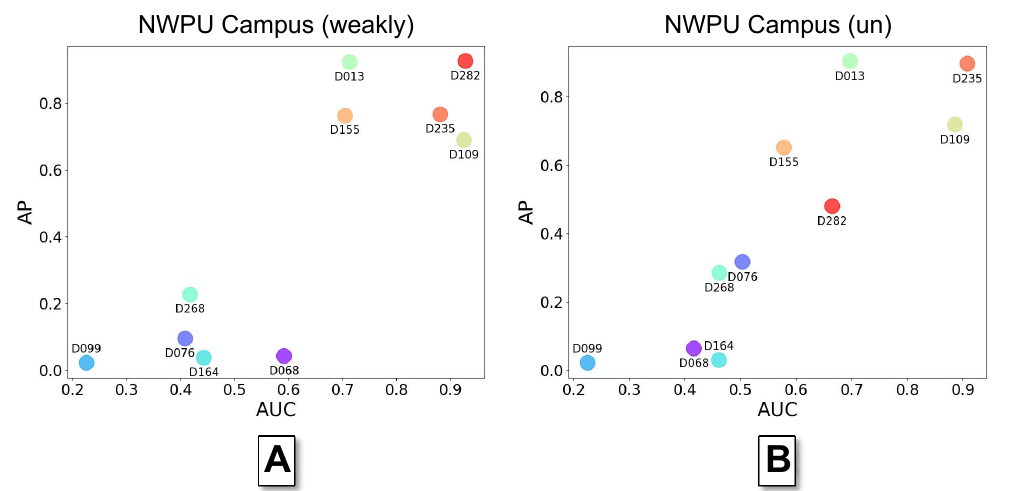}}
	\vspace{-0.6cm}
	\caption{The scatter plot depicts our model's anomaly detection capability in various scenarios (weakly supervised and unsupervised). The horizontal axis represents AUC, while the vertical axis represents AP. Closer proximity to the top-right corner indicates a stronger detection ability of the model. The labels of the scatter points represent the scenario IDs.}
	\label{fig:nwpuc}
	\vspace{-0.25cm}
\end{figure}

\begin{figure}[!t]
	\centering{\includegraphics[width=1\linewidth]{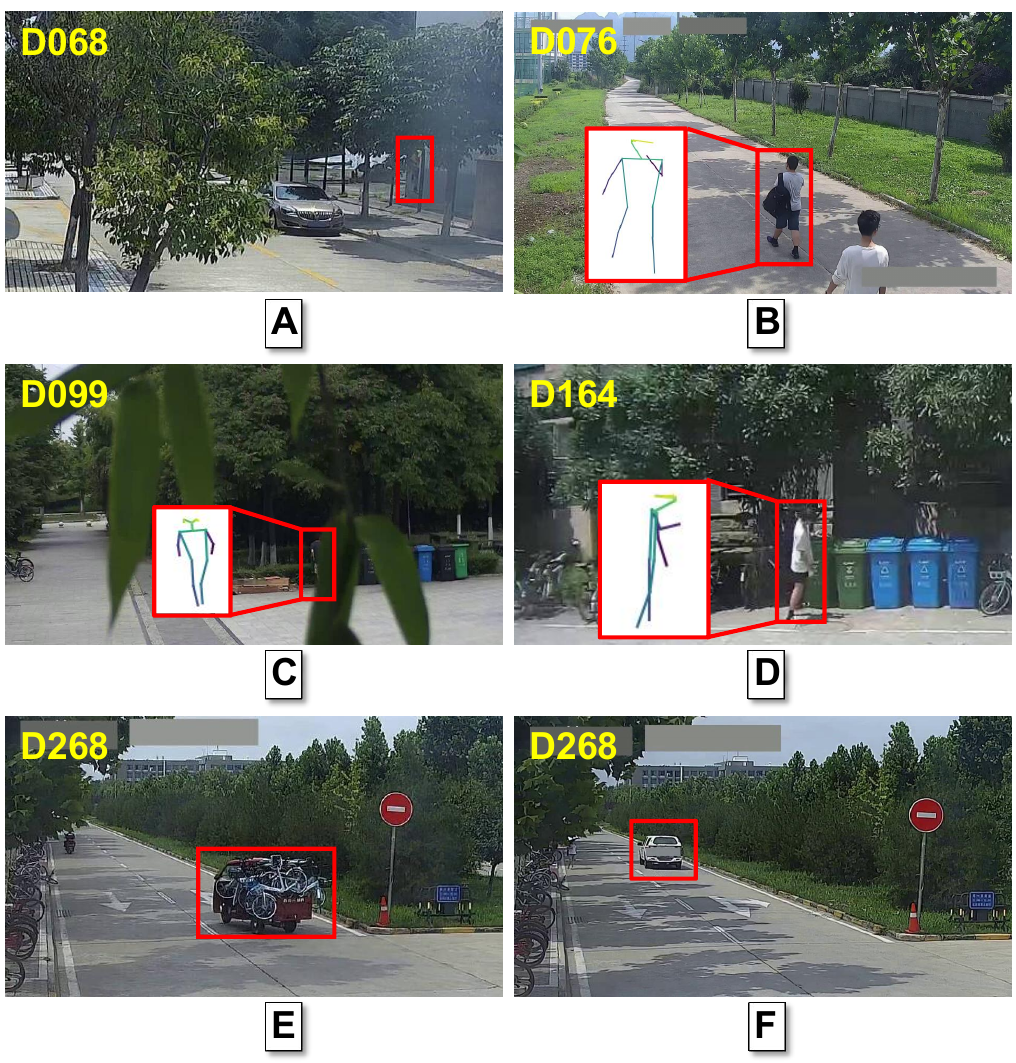}}
	\vspace{-0.5cm}
	\caption{Failure cases on the NWPU Campus dataset. The yellow labels in the top-left corner represent the scenario IDs.}
	\label{fig:failure}
	\vspace{-0.45cm}
\end{figure}

We found that the AP performance of all models (including weakly supervised and unsupervised) was poor on the NWPU Campus dataset. We analyzed all scenes in the dataset using weakly supervised and unsupervised models and visualized the performance of the top five and bottom five scenes in Fig.~\ref{fig:nwpuc}. A detailed analysis of the poorly performing scenes (as shown in Fig.~\ref{fig:failure}) revealed that the anomalies in these scenes often involve severe occlusion, significant ambiguity, and a substantial presence of non-human-related anomalies. These factors lead to the models' inability to effectively detect the anomalies, thereby affecting the AP values.


\subsection{Limitations}
While the DecoAD approach shows promise in addressing the limitations of existing human-related video anomaly detection methods, through the analysis of the NWPU Campus dataset (see Sec.~\ref{sec:lim}), we identified some potential limitations of this approach: 1) in cases where behaviors are highly similar, their semantic distance is minimal, making it difficult for the model to accurately distinguish between them. This difficulty is particularly evident when combined with scene context; 2) in complex scenarios involving occlusion and background distractions, there may be errors in skeleton extraction, such as obtaining only partial skeletons. This incomplete skeleton information may lead to incorrect predictions of anomaly scores because the missing semantic context can mislead the model; 3) when dealing with appearance anomalies, such as improper backpack positioning, the model, based on skeleton data for anomaly detection, is unable to recognize these anomalies; 4) for abnormal behaviors not directly involving humans, such as vehicles violating traffic rules, action-based methods are unable to detect them.

Finally, we found that the FPS (frames per second) of our model is relatively low. We further analyzed the time required for each key step in Table \ref{tab:limitations} and discovered that the time consumed in processing a single video frame is primarily concentrated in the ``Scene-Action Decoupling" part, mainly due to the excessive time overhead of skeleton extraction. As skeleton extraction technology advances, there is potential for further improvement in the FPS of our method.

\begin{table}[!t]
	\centering
	\caption{Detailed average time cost for processing a single video frame. This result was obtained on a PC equipped with an Intel(R) Xeon(R) CPU and an NVIDIA GTX 4090 GPU (with 24G RAM). The experiment was conducted on an SSD set.}
	\vspace{-0.2cm}
	\begin{tabular}{c}
		\hspace{-0.3cm}
		\includegraphics[width=1\linewidth]{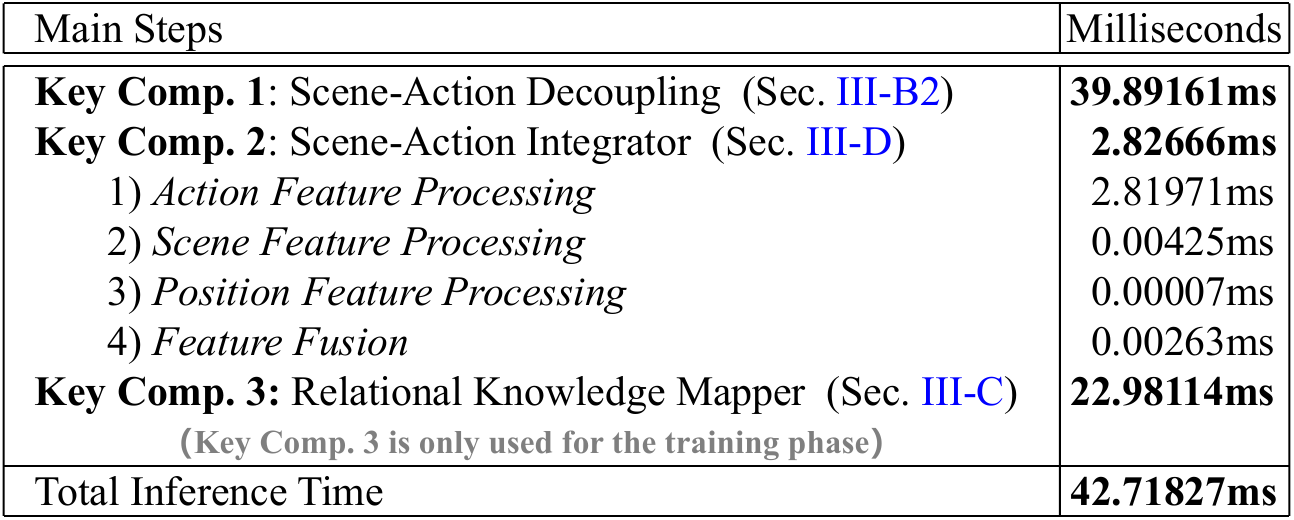}
	\end{tabular}
	\label{tab:limitations}
	\vspace{-0.45cm}
\end{table}

\section{Conclusion}
This study introduces DecoAD, an innovative architecture for detecting anomalies in human-related videos. By employing the concept of ``Scene-Action Interweaving", DecoAD surpasses existing methods in accuracy and robustness to detect context-related anomalies. The proposed methodology involves ``Relation Interweaving'', ``Feature Interweaving'', and ``Uncertainty Refinement'', enabling a comprehensive understanding of the complex relationships between scenes, human actions, and video clips.
Extensive experiments on benchmark datasets demonstrate that DecoAD outperforms state-of-the-art approaches, achieving superior accuracy and robustness.

Future research could focus on challenges such as incomplete skeleton extraction and distinguishing between similar behaviors. Current skeleton extraction technologies often struggle with occlusions or fast movements, which directly impacts the effectiveness of anomaly detection models. Improving algorithms or introducing new technologies could enhance the accuracy of skeleton extraction. Additionally, differentiating behaviors that look similar but have different meanings is crucial. This can be achieved by optimizing feature extraction and classification algorithms, incorporating more contextual information, and utilizing multimodal data to improve model performance. These efforts will enhance the functionality and applicability of the model across a wider range of scenarios.

\bibliographystyle{IEEEtran}
\bibliography{refs}

\end{document}